\documentclass[10pt,conference]{IEEEtran}
\usepackage{enumerate}
\usepackage{graphicx} 

\usepackage{amsmath}
\usepackage{mathtools}
\begin{document}

% paper title
\title{Cognitive State Classification Using Transformed fMRI Data}

\author{
\authorblockN{Hariharan Ramasangu}
\authorblockA{Dept.\ of Electronics and Electrical Engineering, \\
M. S. Ramaiah School of Advanced Studies,\\
Bangalore, India\\
Email: hariharan@msrsas.org} \and
\authorblockN{Neelam Sinha}
\authorblockA{Biomedical Group,\\
International Institute of Information Technology,\\
Bangalore, India\\
Email: neelam.sinha@iiitb.ac.in}

}

\maketitle

\begin{abstract}
One approach, for understanding human brain functioning, is to analyze the changes in the brain while performing cognitive tasks. Towards this, Functional Magnetic Resonance (fMR) images of subjects performing well-defined tasks are widely utilized for task-specific analyses. In this work, we propose a procedure to enable classification between two chosen cognitive tasks, using their respective fMR image sequences. The time series of expert-marked anatomically-mapped relevant voxels are processed and fed as input to the classical Naive Bayesian and SVM classifiers. The processing involves use of random sieve function, phase information in the data transformed using Fourier and Hilbert transformations. This processing results in improved classification, as against using the voxel intensities directly, as illustrated. The novelty of the proposed method lies in utilizing the phase information in the transformed domain, for classifying between the cognitive tasks along with random sieve function chosen with a particular probability distribution. The proposed classification procedure is applied on a publicly available dataset, StarPlus data, with 6 subjects performing the two distinct cognitive tasks of watching either a picture or a sentence. The classification accuracy stands at an average of 65.6\%(using Naive Bayes classifier) and 76.4\%(using SVM classifier) for raw data. The corresponding classification accuracy stands at  96.8\% and 97.5\% for Fourier transformed data. For Hilbert transformed data, it is 93.7\% and 99\%, for 6 subjects, on 2 cognitive tasks.
\end{abstract}

\section{Introduction}
Over the last decade, several aspects of the human brain functioning have been extensively studied using functional Magnetic Resonance Imaging (fMRI). fMRI has become a very important imaging modality in diagnosing, treating and monitoring several brain disorders such as Schizophrenia \cite{schizo}, Alzheimer's disease \cite{AD} etc. The utility of fMR imaging is mainly because the activity of the brain can be studied at sub-second temporal resolution, and sub-mm spatial resolution, allowing detailed studies. Among the several functional aspects that are studied using fMRI, are cortical mapping of cognitive tasks, networks in cognitive tasks, functional disorders, effects of medication therapies, improvement or deterioration in an affected brain. Hence fMRI is used by various groups for studies of neuroscience and psychiatry.

fMR imaging does not use any external contrast agent, instead it exploits the changes in blood oxygen levels that occur in the brain at regions of neural activity. This change in oxygenation, termed Blood Oxygen Level Dependent (BOLD) signal, indicates the level of neural activity in the brain volume, captured at points (voxels) along a 3D grid. In fMR imaging, the subject is continously scanned over a well-designed period of time, during which the subject follows instructions of either executing a well-designed task or resting. Hence  images of the entire brain volume are generated, all along the time-course of the experiment. Thus, if the scanner acquires data every second, and the experiment runs for a course of 100 seconds, one would have 100 volumes of brain data.

The initial works on fMRI analysis focussed on task-specific cortical mapping \cite{friston}, \cite{friston1}, using the Generalized Linear Model approach, where all voxels were processed independently. However, later works  involved studying cognitive states, where the entire brain volume was considered as a single pattern. In their pioneering work on decoding cognitive states \cite{decode}, the authors have elaborated the challenges that arise in this approach. The fraction of the number of relevant voxels is typically very small as compared to the entire number of voxels leading to the challenge of feature selection. Besides the number of available samples for a given cognitive state is far less than the dimension of the feature vectors. Hence the search for the optimal feature set, which could be used with a best-suited classifier has been the focus of several works \cite{Pereira,Ryali}.

\section{Related Work}
Classification experiments on six cognitive tasks, performed by five subjects, using activation maps have been reported in the literature \cite{lee}. The activation maps were used to derive feature vectors, from regions that were marked as consistently and exclusively activated for a given task, during training. The Support Vector Machine (SVM) classifier was used for classification. The authors have reported an average classification accuracy of 74.5\% across the five subjects performing six cognitive tasks. However, the drawback of the proposed method is that it requires a training process where experts identify task-specific neural regions to extract feature vectors. In \cite{yong} the authors have addressed the issue of building classifiers that  detect a particular cognitive state across different subjects using fMR images. This study also uses training data to extract the relevant brain regions. This is followed by utilizing statistical information of the brain regions to form features. The classification performance of this method is validated in a deception fMRI study, using SVM classifier.

The challenges in using the fMR image sequence for classifying the cognitive tasks is that the dimension of the data is several magnitudes larger than the available samples for training. Hence it is important to choose relevant features or transform them to a domain where classification becomes more efficient. Most of the progress in fMRI analysis has been to determine the optimal combination of features and classifiers. 

\begin{figure} [!h]
\begin{center}  
\includegraphics[height=2in,width=3in,angle=0]{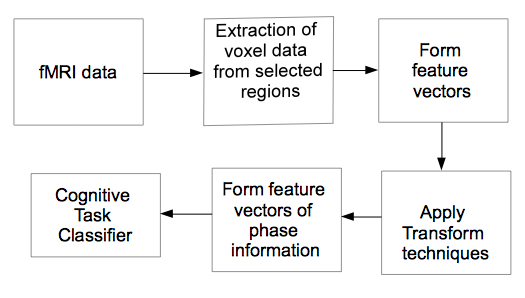}  
\caption{\small \sl Block diagram of the proposed method \label{fig:blk_fig}}  
\end{center}  
\end{figure}

In this paper, we address the issue of transforming the feature set, and compare the classification performance of the transformed feature set as against using the raw data. The proposed work explores if classification of time series corresponding to different cognitive tasks could be improved. In this work attributes of the time series, such as periodicity, harmonics, and phase are exploited. The study in this paper aims to classify between two cognitive tasks using a combination of feature extraction and iterative classification. The block diagram in Fig. \ref{fig:blk_fig} shows the proposed method. Given a time series of fMR images that correspond to the subject performing certain well-designed cognitive tasks, an expert marks out voxels that are relevant for the classification. The time-series of voxel intensities (here, called raw data) along these regions are separated. The raw data is transformed using transformations such as Fourier transform and Hilbert transform \cite{king} to obtain the complex-valued data after applying random sieve function, whose phase information is processed and utilized as features. The transformed data is given as input to Naive Bayesian (NB) and SVM classifiers, to obtain the final classification.

\section{Proposed method}
The proposed method is based on the observation that the fMRI data results in  voxel intensities that exhibit a quasi-periodic pattern. The assumption of quasi-periodicity is made since the same set of cognitive tasks are repeatedly performed. The phase information, in the spectral analysis, holds  key aspects of the features used for classification tasks. Identification of discriminating features from the regions of interest is an important aspect. We have proposed a transformation of fMRI data with random sieve function followed by phase extraction using either Fourier or Hilbert transforms. The utility of classifying data in the transformed domain is motivated by generalizations of analytic signal theory to higher dimensional problems \cite{Bernstein2013}.

Let $X(j)$ be $m$ integers sampled uniformly from the set $\{1,2,\ldots,N\}$.  $\gamma(n)$, where $n\in\{1,2,3,\ldots,N\}$ is defined as follows,

\begin{equation}
\gamma(n) = 
\begin{dcases}
    0, & \exists j \in \{1,2,\ldots,m\}   \ni n=X(j)\\
    1,              & \text{otherwise}
\end{dcases}
\end{equation}

The random sieve function $S_\gamma$ is defined as a mapping from raw voxel data,$f(n)$, to $g(n)$ and $.$ represents coordinate-wise multiplication.:

\begin{equation}
S_\gamma: f(n) \to g(n)
\end{equation}

\begin{equation}
g(n) = f(n) . \gamma(n)
\end{equation}

The Discrete Fourier Transform (DFT) of a sequence $g(n)$ is given by $G(k)$, where $k$ varies from $1$ to $N$,

\begin{equation} 
G(k) = \sum_{n=1}^{n = N}{g(n)}{e^{-i2{\pi}k(n-1)/N}}
\end{equation}
This transform results in the complex-valued spectral decomposition.  Spectral magnitude and phase are the two important features in the
Fourier representation of the signal. Under certain conditions, the given signal can be reconstructed using only the phase information in the Fourier representation. The proposed work utilizes the phase information for classification of the cognitive tasks. If the phase provides an important information about the features, a shift in phase may enhance or corrupt the discriminative nature of the given features. This leads to Hilbert transform which acts as an asymmetric phase shifting operator. The Discrete Hilbert Transform (DHT) has many forms of representation. It is defined using Inverse Discrete Time Fourier Transform (IDTFT),

\begin{equation} 
H(k) = IDTFT\{G(\omega) \sigma_H(\omega)\}
\end{equation}

\begin{equation}
\sigma_H(\omega) = 
\begin{dcases}
    e^{i{\pi}/2}, & \text{if } -\pi < \omega < 0\\
    e^{-i{\pi}/2}, & \text{if } 0 < \omega < \pi\\
    0,              & \text{otherwise}
\end{dcases}
\end{equation}

This transform also results in complex values, where the imaginary part of the transformed data captures the phase information. The raw voxel data has been transformed by random sieve function before mapped to Fourier and Hilbert domains. The phase information has been captured from the transformed data, in both the cases, and used for classification. The phase information is utilized to classify between the different time series that correspond to distinct cognitive tasks and not used to recover the signal.

Naive Bayes and Support Vector Machine classifiers are used for cognitive task classification in this work. The improvement in classification accuracy is clearly due to the use of transformed data. The sequence of voxel data has the spatial order mix-up as spatially-neighbour voxels may not be sequential neighbours. This mix-up and re-ordering of neighbourhood while generating voxel features suggest the existence of asymmetrical phase shifting which is captured by the use of Hilbert transform. The mix-up is important as it might enhance the classification by creating new features by combining different spatial voxels. The idea behind the proposed method is classification of cognition tasks in the transformed domain using phase information instead of utilizing the raw fMRI data. 

The purpose of random sieve function is to reduce the number of data points considered for a good classification performance. Under the assumption that sufficient information is contained in lesser number of voxels, random sieve function is used to reduce the number of voxels in the input data, without creating any structural bias in the selection. The following eight classifier configurations have been compared to bring out the impact of transformation using random sieve function and spectral analysis. $arg(.)$ represents the angle of a given complex number.

\begin{equation}
f(n) \xrightarrow {\text{NB}} C_1
\end{equation}

\begin{equation}
f(n) \xrightarrow {\text{SVM}} C_2
\end{equation}

\begin{equation}
f(n) \xrightarrow {\text{$S_\gamma$}} g(n) \xrightarrow {\text{NB}} C_3
\end{equation}

\begin{equation}
f(n) \xrightarrow {\text{$S_\gamma$}} g(n) \xrightarrow {\text{SVM}} C_4
\end{equation}

\begin{equation}
f(n) \xrightarrow {\text{$S_\gamma$}} g(n) \xrightarrow {\text{arg(DFT)}} arg(G(k)) \xrightarrow {\text{NB}} C_5
\end{equation}

\begin{equation}
f(n) \xrightarrow {\text{$S_\gamma$}} g(n) \xrightarrow {\text{arg(DFT)}} arg(G(k)) \xrightarrow {\text{SVM}} C_6
\end{equation}

\begin{equation}
f(n) \xrightarrow {\text{$S_\gamma$}} g(n) \xrightarrow {\text{arg(DHT)}} arg(H(k)) \xrightarrow {\text{NB}} C_7
\end{equation}

\begin{equation}
f(n) \xrightarrow {\text{$S_\gamma$}} g(n) \xrightarrow {\text{arg(DHT)}} arg(H(k)) \xrightarrow {\text{SVM}} C_8
\end{equation}

The dataset and results are elaborated in the next section. 

\section{Data and Results}
\subsection{Star Plus data}
The dataset, called "StarPlus data", was downloaded from the website  \cite{star}. The experiment consisted of 54 trials. In one trial of the experiment, the subject was sequentially shown a picture and a sentence, and was told to press a button to determine if the picture correctly matched the sentence. The pictures were geometric arrangement of symbols such as $*$, $+$ and/or $\$$. The sentences were descriptions of the shown picture such as: “It is true that the star is above the plus”. The picture was presented first on the first half of the trials, while on the other half, the sentence was shown first. Snapshots of the brain were collected every 0.5 seconds. The data was marked with 25-30 anatomically defined regions referred to as Regions of Activation (ROA). The data consists of a set of trials. Each trial consists of an average of 5000 voxels for one snapshot totalling to about 270000 voxels for a particular trial over the entire span of time in the experiment. For a particular subject, 40 trials of data are collected for each cognition task. Hence, there are 80 trials of data for each subject. This data is stored for 6 subjects in the experiment for analysis.

\begin{enumerate}
\item{The first stimulus (S or P) was presented at the beginning of the trial.}
\item{The stimulus was removed after 4 seconds, replaced by a blank screen}
\item{The second stimulus was presented after 4 seconds. This was presented for 4 seconds, or until the subject pressed the button, whichever came first.}
\item{A rest period of 15 seconds was allowed after the second stimulus was removed from the screen.}
\end{enumerate}

The available StarPlus data has already been preprocessed to remove artefacts due to head motion, signal drift, and other sources. Experts have marked the relevant anatomical regions in the brain that participate in the mentioned cognitive tasks. The voxels only from the ear-marked areas ('CALC', 'LIPL', 'LT', 'LTRIA', 'LOPER', 'LIPS', and 'LDLPFC' - as given in the web-site \cite{star}) are chosen as features for classification.

\begin{table}[!h]
\caption{Average Classifier Performance}
\centering
\begin{tabular}{c c}
\hline\hline
 Classifier Configuration & Correct Classification in \%  \\ [0.5ex] % inserts table %heading
\hline
$C_1$ & 65.6 \\
$C_2$ & 76.4 \\
$C_3$ & 60.8 \\
$C_4$ & 69.1 \\
$C_5$ & 96.8 \\
$C_6$ & 97.5 \\
$C_7$ & 93.7 \\
$C_8$ & 99.0 \\ [1ex]
\hline
\end{tabular}
\label{table:result_avg1}
\end{table}

All simulations have been carried out in MATLAB. The reported trials are limited to single-subject classification. Only activated voxels are considered to reduce the dimension of the input data. The time series data corresponding to each of the cognitive tasks is chosen to create the relevant data sets. For each of the data sets, 40 samples for each of the cognitive tasks is chosen. Leave-one-out cross validation was performed, for example, 79 out of 80 samples constitute the training set, whereas the remaining sample served as the test input. We have tried many values and found $N=14000$ gives the improved performance while defining the random sieve function. The determination of $N$ may be based on cross validation which needs to be investigated. The tabulations in Table \ref{table:result_avg1} compare the classification performance for eight classifier configurations given in the previous section, averaged over all six subjects. Since the Random Sieve Function (RSF) is involved, the experiments have been repeated 50 times and average values are given in Table \ref{table:result_avg1}. The standard deviation lies  between 1 and 2 for these cases.

\begin{table}[!h]
\caption{Confusion matrix (subject 5) - Raw data (NB) - $C_1$}
\centering
\begin{tabular}{c c c}
\hline\hline
Class labels & Class-1 & Class-2  \\ [0.5ex] % inserts table %heading
\hline
Class-1 & 24 & 16 \\
Class-2 & 12 & 28 \\ [1ex]
\hline
\end{tabular}
\label{table:cms5c1}
\end{table}

\begin{table}[!h]
\caption{Confusion matrix (subject 5) - Raw data (SVM) - $C_2$}
\centering
\begin{tabular}{c c c}
\hline\hline
Class labels & Class-1 & Class-2  \\ [0.5ex] % inserts table %heading
\hline
Class-1 & 29 & 11 \\
Class-2 & 9 & 31 \\ [1ex]
\hline
\end{tabular}
\label{table:cms5c2}
\end{table}

\begin{table}[!h]
\caption{Confusion matrix (subject 5) - Raw data + RSF (NB) - $C_3$}
\centering
\begin{tabular}{c c c}
\hline\hline
Class labels & Class-1 & Class-2  \\ [0.5ex] % inserts table %heading
\hline
Class-1 & 20 & 20 \\
Class-2 & 10 & 30 \\ [1ex]
\hline
\end{tabular}
\label{table:cms5c3}
\end{table}

\begin{table}[!h]
\caption{Confusion matrix (subject 5) - Raw data + RSF (SVM) - $C_4$}
\centering
\begin{tabular}{c c c}
\hline\hline
Class labels & Class-1 & Class-2  \\ [0.5ex] % inserts table %heading
\hline
Class-1 & 28 & 12 \\
Class-2 & 9 & 31 \\ [1ex]
\hline
\end{tabular}
\label{table:cms5c4}
\end{table}

\begin{table}[!h]
\caption{Confusion matrix (subject 5) - RSF+DFT (NB) - $C_5$}
\centering
\begin{tabular}{c c c}
\hline\hline
Class labels & Class-1 & Class-2  \\ [0.5ex] % inserts table %heading
\hline
Class-1 & 38 & 2 \\
Class-2 & 2 & 38 \\ [1ex]
\hline
\end{tabular}
\label{table:cms5c5}
\end{table}

\begin{table}[!h]
\caption{Confusion matrix (subject 5) - RSF+DFT (SVM) - $C_6$}
\centering
\begin{tabular}{c c c}
\hline\hline
Class labels & Class-1 & Class-2  \\ [0.5ex] % inserts table %heading
\hline
Class-1 & 39 & 1 \\
Class-2 & 1 & 39 \\ [1ex]
\hline
\end{tabular}
\label{table:cms5c6}
\end{table}

\begin{table}[!h]
\caption{Confusion matrix (subject 5) - RSF+DHT (NB) - $C_7$}
\centering
\begin{tabular}{c c c}
\hline\hline
Class labels & Class-1 & Class-2  \\ [0.5ex] % inserts table %heading
\hline
Class-1 & 35 & 5 \\
Class-2 & 5 & 35 \\ [1ex]
\hline
\end{tabular}
\label{table:cms5c7}
\end{table}

\begin{table}[!h]
\caption{Confusion matrix (subject 5) - RSF+DHT (SVM) - $C_8$}
\centering
\begin{tabular}{c c c}
\hline\hline
Class labels & Class-1 & Class-2  \\ [0.5ex] % inserts table %heading
\hline
Class-1 & 40 & 0 \\
Class-2 & 0 & 40 \\ [1ex]
\hline
\end{tabular}
\label{table:cms5c8}
\end{table}

%
%
%\begin{table}[!h]
%\caption{Confusion matrix (subject 5) - Fourier transformed data}
%\centering
%\begin{tabular}{c c c}
%\hline\hline
%Class labels & Class-1 & Class-2  \\ [0.5ex] % inserts table %heading
%\hline
%Class-1 & 29 & 11 \\
%Class-2 & 13 & 27 \\ [1ex]
%\hline
%\end{tabular}
%\label{table:cm_ftdata5}
%\end{table}
%
%\begin{table}[!h]
%\caption{Confusion matrix (subject 5) - Hilbert transformed data}
%\centering
%\begin{tabular}{c c c}
%\hline\hline
%Class labels & Class-1 & Class-2  \\ [0.5ex] % inserts table %heading
%\hline
%Class-1 & 25 & 15 \\
%Class-2 & 12 & 28 \\ [1ex]
%\hline
%\end{tabular}
%\label{table:cm_htdata5}
%\end{table}
%
%\begin{table}[!h]
%\caption{Confusion matrix (subject 6) - Raw data}
%\centering
%\begin{tabular}{c c c}
%\hline\hline
%Class labels & Class-1 & Class-2  \\ [0.5ex] % inserts table %heading
%\hline
%Class-1 & 30 & 10 \\
%Class-2 & 12 & 28 \\ [1ex]
%\hline
%\end{tabular}
%\label{table:cm_rawdata6}
%\end{table}
%
%\begin{table}[!h]
%\caption{Confusion matrix (subject 6) - Fourier transformed data}
%\centering
%\begin{tabular}{c c c}
%\hline\hline
%Class labels & Class-1 & Class-2  \\ [0.5ex] % inserts table %heading
%\hline
%Class-1 & 30 & 10 \\
%Class-2 & 12 & 28 \\ [1ex]
%\hline
%\end{tabular}
%\label{table:cm_ftdata6}
%\end{table}
%
%\begin{table}[!h]
%\caption{Confusion matrix (subject 6) - Hilbert transformed data}
%\centering
%\begin{tabular}{c c c}
%\hline\hline
%Class labels & Class-1 & Class-2  \\ [0.5ex] % inserts table %heading
%\hline
%Class-1 & 28 & 12 \\
%Class-2 & 10 & 30 \\ [1ex]
%\hline
%\end{tabular}
%\label{table:cm_htdata6}
%\end{table}

\section{Discussion}
Although Fourier and Hilbert Transforms have been applied in several domains \cite{hahn} \cite{alaif} \cite{feldman}, their utility in processing fMRI data for the purpose of classifying between cognitive states, along with random sieve function, is a novel contribution. The applications of these transforms stem from the observation that the cognitive tasks elicit certain periodic patterns. This inherent periodicity in the voxel values can be exploited to distinguish between cognitive tasks. The voxel features are taken as a sequence for classification tasks. Naive Bayesian classifier is a simple classifier for data that is assumed independent. This classifier is proven to be effective in spite of the restrictive assumption of independence, in several applications. The idea behind this classifier is that the posterior is proportional to the prior and that the proportionality factor varies directly with the likelihood. The focus of this paper is to bring out the effect of random sieve function and spectral transformation on the performance of classifiers. The performance of the proposed approach for other transform techniques has to be investigated in the future. The work is concerned with the use of raw data and the transformed data for classification and not the comparison of classification schemes. Classification based on other techniques such as regression coefficients, summary statistic, etc. could also be applied on transformed data.

It is evident, from Table \ref{table:result_avg1}, that the classification accuracy has been improved for the proposed transformation. The average classification accuracy using the raw data, on a Naive Bayesian classifier, stands at 65.6\%. On the other hand, using the transformed data, the classification accuracy obtained on the same Naive Bayesian classifier, is 96.8\% and 93.7\%, respectively for Fourier and Hilbert transformations along with random sieve function. The average classification accuracy using the raw data, on SVM classifier (hard margin), stands at 76.4\%. Outliers in the data would influence the classification boundary in hard-margin SVM. On the other hand, using the transformed data, the classification accuracy obtained on the same SVM classifier, is 97.5\% and 99.0\%, respectively for Fourier and Hilbert transformations along with random sieve function.

 The tabulations in Table \ref{table:cms5c1}, \ref{table:cms5c2}, \ref{table:cms5c3}, \ref{table:cms5c4}, \ref{table:cms5c5}, \ref{table:cms5c6}, \ref{table:cms5c7}, and \ref{table:cms5c8}, show the typical confusion matrix obtained for each of the 8 classifier configuration, for Subject-5 data.

%\begin{figure}  
%\begin{center}  
%%%\includegraphics[height=7in,width=5in,angle=90]{file.eps}  
%\caption{\small \sl Representative confusion matrix for the 3 cases \label{fig:conf_mat}}  
%\end{center}  
%\end{figure}  

The sequence generated from 3D spatial structure implies that there is a mix-up of order when we arrange them as 1D vector. The change in neighbourhood information due to this re-order leads to two cases : adjacent occurrences of voxels do not imply spatial neighbourhood and spatial neighbours may not be closer. The asymmetrical re-ordering of voxel features is an important factor which has not normally been taken into account for classification tasks. By using the transformed data, the proposed approach utilizes this asymmetrical phase-shift explicitly and this leads to improved classification accuracy. The phase components in Fourier domain and the imaginary components of Hilbert transform are given to the classifier. From the classifier point of view, the difference between using the raw data and transformed data is coming from the explicit use of phase information along with random sieve function. The above discussion also opens up the scope of the proposed approach as phase analysis is widely used in communication systems. The future investigations would involve the evaluation of phase algorithms, multivariate methods, other transform techniques, N-fold cross validation, and sieve methods, for fMRI data classification and their theoretical implications. 

\section{Conclusion}
In this paper, we have proposed a random sieve function along with the utility of transforms on voxel intensities, for classifying cognitive states.  The two cognitive tasks considered here are, "sentence viewing" and "picture viewing". Real datasets, StarPlus data, comprising of fMR data obtained from 6 healthy volunteers are used. The experiment consists of transforming the data using random sieve function, Fourier Transform  and Hilbert Transform, prior to using Naive Bayesian and SVM classifiers. The results obtained show an improvement of about 30\% while using Fourier transformed data and about 23\% for Hilbert transformed data, as against using raw voxel intensities.


\begin{thebibliography}{1}



\bibitem{schizo}
Shenton, Martha E., et al., "A review of MRI findings in schizophrenia," \emph{Schizophrenia Research}, vol. 49, no. 1, pp. 1-52, 2001.

\bibitem{AD}
Rombouts, Serge ARB, et al., "Altered resting state networks in mild cognitive impairment and mild Alzheimer's disease: an fMRI study," \emph{Human brain mapping}, vol. 26, no. 4, pp. 231-239, 2005.

\bibitem{friston}
Worsley, Keith J., and Karl J. Friston, "Analysis of fMRI time-series revisited—again," \emph{Neuroimage}, vol. 2, no. 3, pp.173-181, 1995.

\bibitem{friston1}
Friston, K. J., et al., "Event-related fMRI: characterizing differential responses," \emph{Neuroimage}, vol. 7, no. 1, pp. 30-40, 1998. 

\bibitem{decode}
Mitchell, Tom M., et al.,"Learning to decode cognitive states from brain images," \emph{Machine Learning}, vol. 57, no. 1-2, pp. 145-175, 2004.

\bibitem{Pereira}
Pereira, Francisco, Tom Mitchell, and Matthew Botvinick, "Machine learning classifiers and fMRI: a tutorial overview," \emph{Neuroimage}, vol. 45, no. 1, pp. S199-S209, 2009.

\bibitem{Ryali}
Ryali, Srikanth, et al., "Sparse logistic regression for whole-brain classification of fMRI data," \emph{NeuroImage}, vol. 51, no. 2, pp. 752-764, 2010.

\bibitem{lee}
Lee, Jong-Hwan, et al., "Automated classification of fMRI data employing trial-based imagery tasks," \emph{Medical Image Analysis}, vol. 13, no. 3, pp. 392-404, 2009.

%Med Image Anal. 2009 Jun;13(3):392-404. doi: 10.1016/j.media.2009.01.001. Epub 2009 Jan 16.
%Automated classification of fMRI data employing trial-based imagery tasks.
%Lee JH, Marzelli M, Jolesz FA, Yoo SS.

%Yong, F., D. Shen and C. Davatzikos. (2006). "Detecting Cognitive States from fMRI Images by Machine Learning and Multivariate Classification." %Proceedings of the 2006 Conference on Computer Vision and Pattern Recognition Workshop. 17-22 June 2006.
\bibitem{yong}
Fan, Yong, Dinggang Shen, and Christos Davatzikos., "Detecting cognitive states from fmri images by machine learning and multivariate classification," \emph{Proceedings of the 2006 Conference on Computer Vision and Pattern Recognition Workshop}, June 2006.

\bibitem{king}
King, Frederick W.,"Hilbert transforms," Vol. 2. Cambridge, UK: Cambridge University Press, 2009.

\bibitem{Bernstein2013}
Bernstein, Swanhild, et al.,"Generalized Analytic Signals in Image Processing: Comparison, Theory and Applications." TIM Birkhauser, to be published, 2013.


\bibitem{star}StarPlus fMRI data, http://www.cs.cmu.edu {online} Available: http://www.cs.cmu.edu/afs/cs.cmu.edu/project/theo-81/www/ [Accessed: Jan 30, 2014].

\bibitem{hahn} Hahn, Stefan L., "Hilbert transforms in signal processing," Vol. 2. Boston: Artech House, 1996.

\bibitem{alaif} Al-Aifari, Reema, Michel Defrise, and Alexander Katsevich. ,"Asymptotic analysis of the SVD for the truncated Hilbert transform with overlap," \emph{arXiv preprint} arXiv:1312.4727 (2013).

\bibitem{feldman} Feldman, Michael., "Hilbert transform in vibration analysis." \emph{Mechanical Systems and Signal Processing}, vol. 25, no. 3, pp. 735-802, 2011.




\end{thebibliography}
\end{document}